%% file: elsarticle-template-num.tex
\documentclass[preprint,12pt]{elsarticle}

\usepackage{amssymb}
\usepackage{amsmath}

\usepackage{algorithm}
\usepackage{algpseudocode}

\journal{Nuclear Physics B}

\begin{document}

\begin{frontmatter}



\title{Differential pose optimization in descriptor space - Combining Geometric and Photometric Methods for Motion Estimation}


\author[label1]{Andreas L. Teigen} 
\author[label2]{Annette Stahl}
\author[label1]{Rudolf Mester}

\affiliation[label1]{organization={Department of Computer Science, Norwegian University of Science and Technology},
            city={Trondheim},
            postcode={7034}, 
            country={Norway}}

\affiliation[label2]{organization={Department of Engineering Cybernetics Norwegian University of Science and Technology},
            city={Trondheim},
            postcode={7034}, 
            country={Norway}}

\input{Body/00Abstract}




\end{frontmatter}



\input{Body/01Introduction}
\input{Body/02RelatedWork}
\input{Body/03Method}

\input{Body/04ImplementationDetails}
\input{Body/05Experiments}
\input{Body/06Discussion}
\input{Body/07Conclusion}

\bibliographystyle{elsarticle-num} 
\bibliography{bibliography}

\input{Body/Appendix}



\end{document}

%% file: Body/00Abstract.tex
\begin{abstract}
    One of the fundamental problems in computer vision is the two-frame relative pose optimization problem. Primarily, two different kinds of error values are used: photometric error and re-projection error. The selection of error value is usually directly dependent on the selection of feature paradigm, photometric features, or geometric features. It is a trade-off between accuracy, robustness, and the possibility of loop closing. We investigate a third method that combines the strengths of both paradigms into a unified approach. Using densely sampled geometric feature descriptors, we replace the photometric error with a descriptor residual from a dense set of descriptors, thereby enabling the employment of sub-pixel accuracy in differential photometric methods, along with the expressiveness of the geometric feature descriptor. Experiments show that although the proposed strategy is an interesting approach that results in accurate tracking, it ultimately does not outperform pose optimization strategies based on re-projection error despite utilizing more information. We proceed to analyze the underlying reason for this discrepancy and present the hypothesis that the descriptor similarity metric is too slowly varying and does not necessarily correspond strictly to keypoint placement accuracy.
\end{abstract}

%% file: Body/01Introduction.tex
\section{Introduction}
\begin{figure*}
  \includegraphics[width=\textwidth,height=0.3\textwidth]{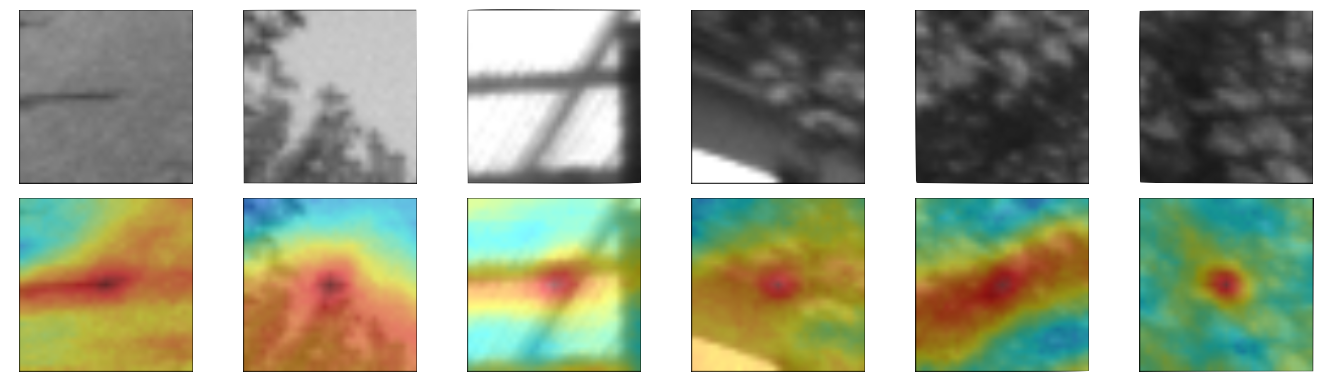}
  \caption{Descriptor self-similarity (Section \ref{sec:self-similarity}). The images on the first row are all image patches centered on different ORB features. The images on the second row display an overlaid heatmap of descriptor similarity for each pixel, measured as the Hamming distance between the binary descriptors of the keypoint in the center of the patch and those of each pixel in the surrounding area. 
  Warmer colors indicate a higher similarity.}
  \label{fig:descriptor_similarity}
\end{figure*}

Optimizing the relative rotation $\mathbf{R}$ and translation $\mathbf{t}$ between two images is one of the fundamental problems of computer vision, and it has been well studied \cite{cadena2016past}. It typically involves establishing an initial feature correspondence set of landmarks visible in both images, applying a minimal solver to estimate an initial pose estimate \cite{nister2004efficient,longuet1981computer,Hruby_2022_CVPR}, usually implemented in conjunction with a \emph{Random Sampling Consensus} (RANSAC) algorithm to remove outliers. To refine this pose and feature placement some sort of iterative, nonlinear optimization scheme is needed. The loss function for these schemes invariably reflects the penalty of shifting the initial corresponding point $\mathbf{y}_k$ to some local point $\mathbf{y}_k + \mathbf{v_k}$ on the epipolar line $\mathbf{r}_k$ such that:

\begin{equation}
    (\mathbf{y}_k^\top + \mathbf{v}_k^\top, \; 1) \; \mathbf{r}_k = 0
\end{equation}

Deciding how this loss is defined and, consequently, which point on the epipolar line is best, is one of the biggest design decisions in classical \emph{Visual Odometry} (VO) and \emph{Simultaneous Localization And Mapping} (SLAM) systems and has resulted in the competing paradigms of the geometric methods and the photometric methods.


Given two images $I_A$ and $I_B$ and two corresponding points $\mathbf{x}_k$ and $\mathbf{y}_k$, \textbf{Geometric methods} forgo using image information, and instead focus solely on minimizing the shortest geometric distance $\mathbf{v}_k$ between the observed point $\mathbf{y}_k$ and the epipolar line $\mathbf{r}_k$.




\begin{equation}
    \text{Loss} = H(\mathbf{v}_k) = ||\mathbf{v}_k||
\end{equation}

This loss function is a paraboloid with a circular cross-section centered on $\mathbf{y}_k$.

\textbf{Photometric methods}, on the other hand, define the loss as the residual between the pixel intensities $P(\mathbf{x}_k)$ in image $I_A$ and $P(\mathbf{y}_k + \mathbf{v}_k)$ in image $I_B$ where $\mathbf{v}_k$ is the delta position between the initial correspondence $\mathbf{y}$ and the point on the point on the epipolar line with the lowest residual pixel intensity.

\begin{equation}
    \text{Loss} = H(P(\mathbf{x}_k), \tilde{P}(\mathbf{y}_k + \mathbf{v}_k)) = |P(\mathbf{x}_k) - \tilde{P}(\mathbf{y}_k + \mathbf{v}_k)|
    \label{eq:photometricLoss}
\end{equation}


Here, $\tilde{P}$ represents a local smoothing of the pixel intensities, allowing differential methods to be utilized. This loss function is largely dependent on the image content and can take on any shape. 

In this paper, we explore a third option, utilizing the descriptors of geometric methods, which are normally used only to establish initial point correspondence, in place of pixel values from photometric methods, to enable differential matching in descriptor space $D(\cdot)$. This enables the use of geometric point descriptors, which are robust to illumination changes, rotational changes, and significant viewpoint changes, while retaining the high pose and point positioning accuracy of photometric methods that result from actively utilizing image information during optimization. This results in the following loss function:

\begin{equation}
    \text{Loss} = D(Q(\mathbf{x}_k), \tilde{D}(\mathbf{y}_k + \mathbf{v}_k)) = |D(\mathbf{x}_k) - \tilde{D}(\mathbf{y}_k + \mathbf{v}_k)|
\end{equation}


This formulation is intentionally similar to Equation \ref{eq:photometricLoss}, in that it can be used as a drop-in replacement in all photometric methods without requiring any additional algorithm modifications, making it universally applicable to these methods.

We first demonstrate that it is possible to change the photometric residual $H(P(\mathbf{x}_k), \tilde{P}(\mathbf{y}_k + \mathbf{v}_k))$ for the descriptor residual $H(D(\mathbf{x}_k), \tilde{D}(\mathbf{y}_k + \mathbf{v}_k))$ by showing that the descriptor space is a smooth function that can be made locally differentiable (Section \ref{sec:self-similarity} \& Figure \ref{fig:descriptor_similarity}). Then, we choose a differential photometric approach, but exchange the photometric residual with the descriptor residual, showing that it results in better results in more demanding conditions (Section \ref{sec:ClassicJet}).

%% file: Body/02RelatedWork.tex
\section{Related work}


Estimating the relative pose between images has a long history, with roots dating back to the 1980s, during the advent of photometric methods using optical flow \cite{horn1981determining,slesareva2005optic,brox2004high,farneback2003two} and photometric feature extractors and matching \cite{lucas1981iterative,shi1994good}. It is widely used today, even in modern SLAM systems \cite{engel2017direct,engel2014lsd}, and is generally considered highly accurate under favorable conditions due to differential point placement methods, providing sub-pixel accuracy. Geometric features \cite{lowe1999object,bay2006surf,rublee2011orb} were introduced later as a way to increase robustness for greater image differences in scale, rotation, lighting, and viewing angles. This opened up for the estimation of the relative pose of a much larger array of images, which led to the introduction of the \emph{photo tourism} methods, which use images from random collections of images from the Web \cite{agarwal2011building,snavely2006photo,snavely2008modeling}.

Both geometric and photometric methods are still used actively today, maybe most prominently in two of the most successful VO/SLAM systems: ORB-SLAM and DSO\cite{mur2015orb,engel2017direct}. ORB-SLAM is the more citetd of the two, one reason for which is its robustness to various scenes and the lack of camera requirements. DSO has more accurate camera tracking, but it requires significantly more from the camera, including knowing the shutter speeds of all images and lens vignetting. These differences largely stem from feature selection and the accompanying matching and frame-to-frame pose optimization strategies. Through this work, we investigate the possibility of developing a third alternative that utilizes the robust feature descriptors of the geometric methods to generate an alternative residual to the pixel intensity residual of the photometric method. Potentially allowing higher accuracies with greater robustness and more lenient camera requirements.

\subsection{Dense descriptors}
Swapping out the pixel intensity residual with a descriptor residual requires the computation of a locally dense set of descriptors. Other works that have experimented with dense descriptors include \cite{tuytelaars2010dense}, which produces a grid with regularly spaced descriptors so that every pixel in the image is covered within the radius of at least one descriptor. They argue that the spatial relations between these descriptors could be a helpful aid in tasks like object detection. Other more recent methods introduce the use of convolutional neural networks \cite{o2015introduction} to generate descriptors for every pixel and then perform dense matching across images \cite{dusmanu2019d2,truong2021learning,revaud2019r2d2}. A more recent method of matching image regions is with machine learning approaches \cite{dusmanu2019d2,sarlin2021back,sarlin2020superglue}. D2-Net \cite{dusmanu2019d2} calculates a dense set of descriptors using a Convolutional Neural Network, and the authors state that the detection stage is actually a hindrance to the matching process and that "approaches that forego the detection stage and instead densely extract descriptors perform much better in challenging conditions. Yet, this gain in robustness comes at the price of higher matching times and memory consumption." This also highlights the opportunity to intelligently identify the correct placement of keypoints, utilizing image information, to improve motion estimation. 

The method that most closely resembles our idea is \cite{liu2010sift}, which utilizes a pixel-wise dense set of SIFT descriptor residuals instead of the photometric residual to perform optical flow. However, this process is computationally demanding and cannot be achieved in a real-time VO/SLAM system. We seek to achieve this in a sparse manner, focusing on accurately placing salient keypoints while only computing locally dense descriptors around them. We also use the more efficient ORB descriptor \cite{rublee2011orb} to reduce the computational demand of the algorithm.

Although any algorithm originally designed with the photometric residual as part of the loss function can be used \cite{lucas1981iterative,delaunoy2014photometric,liu2010sift,bradler2017joint,alismail2016photometric}, we choose to base our work on \cite{bradler2017joint}, which presents a very compact and efficient method to exploit both the epipolar geometry and the local photometric residuals.

%% file: Body/03Method.tex
\section{Method}

\subsection{Descriptor similarity as an alternative to photometric residual.}

The postulate that gives credence to our hypothesis is that descriptors are spatially differentiable functions. This is what allows the descriptor similarity between points to be used instead of the photometric residual. In this section, we show that this postulate holds by checking the self-similarity of the descriptor in an image and that it will also lead to valid results when considering matches between images (\emph{cross-similarity}).


\subsubsection{Self-similarity (Similarity across descriptors in the same image):}
\label{sec:self-similarity}

In order to demonstrate that descriptors are differentiable over an image, a simple experiment is performed: Calculate the descriptors $D(\mathbf{x})$ for all eligible pixels $ \mathbf{x}$ in image $\mathcal{I}_A$, and identify a single salient keypoint $ \mathbf{x}_k$ from that image. The descriptor of that point $D(\mathbf{x}_k)$ is retrieved from the set of dense descriptors $D(\mathbf{x})$ and the descriptor difference $z_k$ is calculated between $D(\mathbf{x}_k)$ and the set of dense descriptors $D(\mathbf{x})$ by determining the descriptor distance $z_k = H(D(\mathbf{x}), D(\mathbf{x}_k))$. This is our self-similarity metric. The scalar descriptor similarity can be visualized as a heat map over the original image. The result of doing this for several different keypoints $\mathbf{x}_k$ from the KITTI dataset is shown in Figure \ref{fig:descriptor_similarity}. For this experiment, we used the ORB detector with accompanying binary descriptors and the Hamming distance as the descriptor difference metric.

The figure shows, as expected, that ORB feature description is a smoothly varying function across the image patch centered on $\mathbf{x}_k$. From the example images, it is also clear that the descriptor similarity function is not nicely formed in a circular shape around the center pixel as it is assumed when using the re-projection error, but rather that it can take on a variety of different and abstract shapes, even retaining high values over a significant portion of the image. 
Another point of note is that although the ORB detector is designed to detect corners in the image, experiments showed several examples where the local region around a keypoint where the descriptor residual followed the shape of an edge or even a plane rather than being confined to only the corner that the algorithm detected. Examples of both of these can be seen in Figure \ref{fig:descriptor_similarity}. This means that although the point detection algorithm is designed to detect salient corners in an image, it might not necessarily be a salient point in descriptor space.

\subsubsection{Cross-similarity (Similarity across descriptors from different images):}
\label{sec:cross-similarity}
To show that it still makes sense to talk about the distances between a descriptor in image $\mathcal{I}_A$ with a group of descriptors in $\mathcal{I}_B$, we also perform a cross-similarity experiment.

First, $N$ keypoints are detected in $\mathcal{I}_A$, and $M$ keypoints are detected in $\mathcal{I}_B$, then normal descriptor-based matching is performed between the two sets of points. Next we select a point $ \mathbf{x}_k$ in $\mathcal{I}_A$ which has a confirmed match $\mathbf{y}_k$ in $\mathcal{I}_B$ and calculate the distance between the descriptor $D(\mathbf{x}_k)$ to the descriptors of all pixels in a local region around $\mathbf{y}_k$. The correctness of the matches is manually checked for this experiment.

The results from the cross-similarity test are largely similar to those from the self-similarity test, with the only difference being the absence of a very pronounced peak at the center of the patch. This is as expected, as there is usually no pixel-perfect match from one image to the next due to a shift in viewing angle.

That being said, when comparing the descriptor of a key-point $\mathbf{x}_k$ in $\mathcal{I}_A$ with the descriptors of the pixels in a local neighborhood around the matched key-point $\mathbf{y}_k$ in $\mathcal{I}_B$, we found that a perfect descriptor $z_k = 0$ match with $\mathbf{x}_k$ could sometimes be found within a radius of 3-5 pixels of $\mathbf{y}_k$ in $\mathcal{I}_B$, given a short baseline between the images. This discovery shows that keypoint detection algorithms based on saliency (at least the ORB detector) can have a significant descriptor distance between an actual match and the ideal pixel match. This also supports the statement in \cite{dusmanu2019d2}, where the authors claim that the detection process limits the matching accuracy of points in an image. 

\subsection{Locally differentiable descriptor residual}
\label{sec:descriptorResidual}

Instead of using the photometric residual, we wish to use the Hamming match residual

\begin{equation}
    Q_k(\mathbf{v}_k) = H(\mathbf{d}(\mathbf{x}_k), \mathbf{d}(\mathbf{y}_k + \mathbf{v}_k))
\end{equation}

between keypoint descriptors $\mathbf{x}_k$ in image $I_A$ and the descriptors for the pixels around the corresponding keypoint $\mathbf{y}_k$ in image $I_B$. A local paraboloid $\mathbf{A}_k$ is fitted for each point $\mathbf{y}_k$. The details of this quadratic fitting are given in the Appendix. Resulting in 

\begin{align}
\begin{split}
    Q_k(\mathbf{v}_k) &\approx \tilde{Q}_k(\mathbf{v}_k) =
    \begin{pmatrix}
        \mathbf{y}_k^T + \mathbf{v}_{k}^T, & 1
    \end{pmatrix} 
    \mathbf{A}_k
    \begin{pmatrix}
        \mathbf{y}_k + \mathbf{v}_{k}\\
        1
    \end{pmatrix}
    \label{eq:quadratic_form}
\end{split}
\end{align}

Since $\mathbf{y}_k$ is the initial estimate of the keypoint location, and consequently a constant, the expression can be further rewritten with respect to the only variable $\mathbf{v}_k$:

\begin{align}
    \begin{split}
        \tilde{Q}_k &=
        \mathbf{v}_k^T
        \begin{pmatrix}
            1 & 0 & 0\\
            0 & 1 & 0
        \end{pmatrix}
        \mathbf{A}_k
            \begin{pmatrix}
            1 & 0 & 0\\
            0 & 1 & 0
        \end{pmatrix}^T
        \mathbf{v}_k \\
        &+ 2 \mathbf{v}_k^T 
        \begin{pmatrix}
            1 & 0 & 0\\
            0 & 1 & 0
        \end{pmatrix}
        \mathbf{A}_k
        \begin{pmatrix}
            \mathbf{y}_k \\
            1
        \end{pmatrix}
        +
        \begin{pmatrix}
            \mathbf{y}_k^T, & 1
        \end{pmatrix}
        \mathbf{A}_k 
        \begin{pmatrix}
            \mathbf{y}_k\\
            1
        \end{pmatrix}\\
        &=
        \mathbf{v}_k^T \mathbf{A'}_k \mathbf{v}_k + 2 \mathbf{v}_k^T \mathbf{b}_k + c_k
        \label{eq:quadratic_form_split}
    \end{split}
\end{align}

with:

\begin{align}
    \begin{split}
        \mathbf{A'}_k &= 
        \begin{pmatrix}
            1 & 0 & 0\\
            0 & 1 & 0
        \end{pmatrix}
        \mathbf{A}_k
            \begin{pmatrix}
            1 & 0 & 0\\
            0 & 1 & 0
        \end{pmatrix}^T\\
        \mathbf{b}_k &= 
        \begin{pmatrix}
            1 & 0 & 0\\
            0 & 1 & 0
        \end{pmatrix}
        \mathbf{A}_k
        \begin{pmatrix}
            \mathbf{y}_k \\
            1
        \end{pmatrix}\\
        c_k &= 
        \begin{pmatrix}
            \mathbf{y}_k^T, & 1
        \end{pmatrix}
        \mathbf{A}_k
        \begin{pmatrix}
            \mathbf{y}_k\\
            1
        \end{pmatrix}
    \end{split}
\end{align}

Resulting in the same starting point as in equation \ref{eq:LucasKanadeMain}. Enabling this formulation using descriptor residual instead of the traditional photometric residual.

The paraboloids are generated based on samples of a few pixels in a local neighborhood around the current operating point $\mathbf{y}_k$.
Consequently, the paraboloids are only valid within a small radius around that point. In the experiments, we generally sampled in a square centered on $\mathbf{y}_k$ with  $\|\mathbf{v}_k\| \leq 7$ pixels.

\section{Joint Epipolar Tracking (JET): A Detailed Mathematical and Conceptual Description}
\label{sec:ClassicJet}

Joint Epipolar Tracking (JET) \cite{bradler2017joint} is a sparse direct visual odometry method that jointly estimates the unscaled relative pose between two images and the sub-pixel accurate feature correspondences. Unlike traditional methods that separate feature matching and pose estimation, JET combines these tasks into a single unified optimization problem. This integration increases robustness and accuracy, particularly in challenging scenarios with limited texture or degraded features.

\subsection{Classical Lucas-Kanade Tracking}
The Lucas-Kanade (LK) method \cite{lucas1981iterative} tracks a keypoint $\mathbf{x}_k$ from a reference image $I_A$ to a target image $I_B$ by minimizing the photometric difference within a small window around the point. The residual is computed as:
\begin{equation}
Q_k(\mathbf{v}_k) = \sum_{\mathbf{x}} W[\mathbf{x}-\mathbf{x}_k]   (I_A[\mathbf{x}] - I_B[\mathbf{y}_k + \mathbf{v}_k])^2,
\end{equation}
where $\mathbf{v}_k$ is the displacement vector and $W$ is a spatial weighting kernel.

Given a point $\mathbf{x}_k$ in image $I_A$ and a current estimate of its corresponding point $\mathbf{y}_k$ in image $I_B$, we can compute not only the photometric residual but also a local approximation of this residual around $\mathbf{y}_k$. This is accomplished using a second-order Taylor expansion, which leads to a locally quadratic model:
\begin{equation}
\tilde{Q}_k(\mathbf{v}_k) = \mathbf{v}_k^\top \mathbf{A'}_k \mathbf{v}_k + 2 \mathbf{v}_k^\top \mathbf{b}_k + c_k,
\label{eq:LucasKanadeMain}
\end{equation}
with:
\begin{align}
\mathbf{A'}_k &= \sum_{\mathbf{x}} W[\mathbf{x}-\mathbf{x}_k] \nabla I_B[\mathbf{y}_k] \nabla I_B[\mathbf{y}_k]^\top, \\
\mathbf{b}_k &= -\sum_{\mathbf{x}} W[\mathbf{x}-\mathbf{x}_k] \nabla I_B[\mathbf{y}_k](I_A[\mathbf{x}] - I_B[\mathbf{y}_k]), \\
c_k &= \sum_{\mathbf{x}} W[\mathbf{x}-\mathbf{x}_k](I_A[\mathbf{x}] - I_B[\mathbf{y}_k])^2.
\end{align}
This local paraboloid provides a smooth, differentiable surface on which optimization can be efficiently performed.

\subsection{Epipolar Constrained Tracking}
The epipolar geometry, given by the translation vector $\mathbf{t}$, the rotation matrix $\mathbf{R}$ and the intrinsic camera matrix $\mathbf{K}$, can be represented by the fundamental matrix:

\begin{equation}
    \mathbf{F} = \mathbf{K}^{-\top} \; [\mathbf{t}]_x \; \mathbf{R} \; \mathbf{K}^{-1}
\end{equation}

where $[\cdot]_x$ is the skew symmetric matrix.

If the fundamental matrix is known, the search for the point in $I_B$ corresponding to $\mathbf{x}_k$ from $I_A$ can be restricted to the epipolar line. This gives the constraint:
\begin{equation}
(\mathbf{y}_k^\top + \mathbf{v}_k^\top, 1) \; \mathbf{F} \; \begin{pmatrix}\mathbf{x}_k \\ 1\end{pmatrix} = 0.
\end{equation}

Treating $\mathbf{y}_k$ as constant and treating $\mathbf{v}_k$ as the variable lets us reformulate the equation:

\begin{align}
    \begin{split}
        \begin{pmatrix}
        \mathbf{y}_k^T + \mathbf{v}_k^T, & 1
        \end{pmatrix}
        \mathbf{F}
        \begin{pmatrix}
            \mathbf{x}_k\\
            1
        \end{pmatrix}
        &= 0,\\
        \mathbf{v}_k^T
        \underbrace{
        \begin{pmatrix}
            1 & 0 & 0\\
            0 & 1 & 0
        \end{pmatrix}
        \mathbf{F}
        }_{\mathbf{F}'}
        \begin{pmatrix}
            \mathbf{x}_k\\
            1
        \end{pmatrix}
        &= -
        \begin{pmatrix}
            \mathbf{y}_k, & 1
        \end{pmatrix}
        \mathbf{F}
        \begin{pmatrix}
            \mathbf{x}_k\\
            1
        \end{pmatrix}
        \label{eq:epipolar_constraint}
    \end{split}
\end{align}

Constraining optimization of the matching residual (equation $\ref{eq:LucasKanadeMain}$) along the epipolar line (equation \ref{eq:epipolar_constraint}) gives results in the formulation of the quadratic optimization problem.

\begin{equation}
\mathbf{v}_{k,opt} := \underset{\mathbf{v}_k}{\arg\min}~ \tilde{Q}_k(\mathbf{v}_k) \quad \text{subject to} \quad (\mathbf{y}_k^\top + \mathbf{v}_k^\top, 1) \; F \begin{pmatrix}\mathbf{x}_k \\ 1\end{pmatrix} = 0,
\end{equation}

This can be solved by first representing it as a Lagrangian equation:

\begin{align}
\mathcal{L}_k(\mathbf{v}_k, \lambda) = \tilde{Q}_k(\mathbf{v}_k) + 2 \lambda \; (\mathbf{y}_k^\top + \mathbf{v}_k^\top, 1) \; \mathbf{F} \; \begin{pmatrix}\mathbf{x}_k \\ 1\end{pmatrix} .
\label{eq:lagrangian}
\end{align}

where $\lambda$ is the Lagrangian multiplier.

The optimal displacement $\mathbf{v}_k$ is found where the partial derivatives of the Lagrangian equate to zero:

\begin{align}
    \frac{1}{2} \frac{\partial \mathcal{L}_k}{\partial \mathbf{v}_{k}}(\mathbf{v}_{k,opt}, \lambda) &= \frac{\partial \tilde{Q}(\mathbf{v}_{k,opt})}{\partial \mathbf{v}_{k}} + \lambda \; \mathbf{F}' 
    \begin{pmatrix}
            \mathbf{x}_k\\
            1
    \end{pmatrix} = 0 \label{eq:lagrangian_partial_v}\\
    \frac{1}{2} \frac{\partial \mathcal{L}_k}{\partial \lambda}(\mathbf{v}_{k,opt}, \lambda) &= 
    \begin{pmatrix}
        \mathbf{y}_k^T + \mathbf{v}_{k,opt}^T, \; 1
    \end{pmatrix}
    \mathbf{F}   
    \begin{pmatrix}
            \mathbf{x}_k\\
            1
    \end{pmatrix} = 0
    \label{eq:lagrangian_partial_lambda}
\end{align}

Substituting $\tilde{Q}_k(\mathbf{v}_k)$ with the quadratic loss function of equation \ref{eq:quadratic_form_split} into equations \ref{eq:lagrangian}, \ref{eq:lagrangian_partial_v} and \ref{eq:lagrangian_partial_lambda} gives us the linear equation system:

\begin{align}
    \begin{split}
    \begin{pmatrix}
        \mathbf{A}'_k & \mathbf{F}'
        \begin{pmatrix}
            \mathbf{x}_k\\
            1
        \end{pmatrix}\\
        \left( \mathbf{F}'
        \begin{pmatrix}
            \mathbf{x}_k\\
            1
        \end{pmatrix}
        \right)^T & 0
    \end{pmatrix}
    \begin{pmatrix}
        \mathbf{v}_{k,opt}\\
        \lambda
    \end{pmatrix}
    \; = \;
    \begin{pmatrix}
        -\mathbf{b}_k\\
        -
        \begin{pmatrix}
            \mathbf{y}_k, & 1
        \end{pmatrix}
        \mathbf{F}
        \begin{pmatrix}
            \mathbf{x}_k\\
            1
        \end{pmatrix}
    \end{pmatrix}
    \end{split}
    \label{eq:epipolar_constrained_tracking}
\end{align}

Matrix $ \mathbf{A}'$ is inherently symmetric, meaning that the entire matrix on the left-hand side of the equation is also symmetric, which means that there is a closed-form solution to this equation system. 

Defining

\begin{align*}
     \mathbf{A}'_k &= 
    \begin{pmatrix}
        a_{11} & a_{12}\\
        a_{12} & a_{22}
    \end{pmatrix},
    && \mathbf{F}'
    \begin{pmatrix}
         \mathbf{x}_k\\
        1
    \end{pmatrix} =
    \begin{pmatrix}
        g_1\\
        g_2
    \end{pmatrix},
    && \mathbf{b}_k = -
    \begin{pmatrix}
        b_1\\
        b_2
    \end{pmatrix},
    &&  \mathbf{v}_{k,opt} =
    \begin{pmatrix}
        v_1\\
        v_2
    \end{pmatrix}
\end{align*}
\begin{align*}
    h &= -
        \begin{pmatrix}
             \mathbf{y}_k, & 1
        \end{pmatrix}
         \mathbf{F}
        \begin{pmatrix}
             \mathbf{x}_k\\
            1
        \end{pmatrix}
\end{align*}

then the closed-form equation system is given as:

\begin{align}
\begin{split}
    v_{1} &= \frac{g_1 g_2 b_2 - a_{22} g_1 h - b_1 g_2^2 + a_{12} g_2 h}{2a_{12} g_1 g_2 - a_{22}g_1^2 - a_11 g_2^2}\\
    v_{2} &= \frac{h}{g_2} - \frac{g_1}{g_2}v_1
\end{split}
\end{align}

\subsection{Joint Epipolar Tracking:}

Since $\mathbf{v}_{k}$ is now defined as the point where the epipolar line intersects the lowest point on the paraboloid, the variable of equation \ref{eq:epipolar_constrained_tracking} can be changed from the displacement $\mathbf{v}_{k,opt}$ to the motion parameters $\mathbf{p} \in \mathbf{SE(3)}$ given the image information $Q(\mathbf{y}_k + \mathbf{v}_{k,opt})$, letting us write:

\begin{equation}
    \mathbf{v}_{k,opt} = \mathbf{f}_k(\mathbf{A}', \mathbf{b}_k, \mathbf{x}_k, \mathbf{y}_k, \mathbf{K}, \mathbf{p}) = \mathbf{f}_k(\mathbf{p})
    \label{eq:v_opt_dependency}
\end{equation}

This means that the problem can be rewritten as dependent only on the motion parameters $ \mathbf{p}$.

Using this substitution of variables, both the epipolar geometry and the point location on the epipolar line can be optimized jointly by considering $\mathbf{p}$ instead of $\mathbf{v}_{k,opt}$ as an optimization parameter in equation \ref{eq:lagrangian}:

\begin{align}
    \begin{split}
        \mathcal{L}_k(\mathbf{p}, \lambda) &= \mathcal{L}_k(\mathbf{v}_k = \mathbf{f}_k(\mathbf{p}))\\
        &= \mathbf{f}_k^T(\mathbf{p}) \mathbf{A}'_k   \mathbf{f}_k(\mathbf{p}) + 2   \mathbf{f}_k^T(\mathbf{p}) \mathbf{b}_k + c_k\\
        &+ 2 \lambda \left(
        \begin{pmatrix}
            \mathbf{y}_k^T + \mathbf{f}_k^T(\mathbf{p}), & 1
        \end{pmatrix}
          \mathbf{F}(\mathbf{p})   
        \begin{pmatrix}
            \mathbf{x}_k\\
            1
        \end{pmatrix} \right)
    \end{split}
    \label{eq:kpt_optimization_motion_parameters}
\end{align}


This is done for all correspondence sets simultaneously, and the summed loss function is as follows:

\begin{equation}
    \mathcal{L}(\mathbf{p}, \lambda) = \frac{1}{N}\sum_{k=1}^N \mathcal{L}_k(\mathbf{p}, \lambda)
\end{equation}

Which gives us the final loss function for the non-linear optimization algorithm with the motion parameters $\mathbf{p}$ as the optimization parameters. 

This results in an incredibly dense formulation of 2-view geometry while utilizing more information than more traditional bundle adjustment methods.

This method is applied iteratively, alternating between calculating the optimal pose and the optimal step $\mathbf{v}_{k,opt}$ and moving $\mathbf{y}_k$ a small step in the direction of $\mathbf{v}_{k,opt}$. The steps should be kept small, since the paraboloid created from the similarities of the descriptors is only valid in a small radius around $\mathbf{v}_{k,opt}$. But since the pose initialization is generally pretty good and $\mathbf{v}_{k,opt}$ tends to be small, not many iterations are required. 
\input{figures/algorithm}

%% file: figures/algorithm.tex
\newlength\myindent
\setlength\myindent{2em}
\newcommand\bindent{%
  \begingroup
  \setlength{\itemindent}{\myindent}
  \addtolength{\algorithmicindent}{\myindent}
}
\newcommand\eindent{\endgroup}

\begin{algorithm}
\caption{D-JET}
\begin{algorithmic}[1]
\State Given:
\State\hspace{\algorithmicindent} Initial correspondences ($\mathbf{x}_k$, $\mathbf{y}_k$)
\State\hspace{\algorithmicindent} Initial pose estimate $\mathbf{p} = \mathbf{p}_0$
\State\hspace{\algorithmicindent} Convergence threshold $\epsilon$ and max iterations

\While{$||\mathbf{p} - \mathbf{p}_{prev}|| > \epsilon$}
    \State $\mathbf{p}_{prev}$ := $\mathbf{p}$
    \State \# Step 1: Optimize point positions along epipolar lines
    \For{each correspondence ($\mathbf{x}_k$, $\mathbf{y}_k$)}
        \State Update local patch residual centered on ($\mathbf{y}_k + \mathbf{v}_{k,opt,prev}$), fit $\mathbf{A'}_k$, $\mathbf{b}_k$, $c_k$
        \State $\mathbf{v}_{k,opt}$ := $\underset{\mathbf{v}_k}{\arg\min}~ \tilde{Q}_k(\mathbf{v}_k)$ subject to $\quad (\mathbf{y}_k^\top + \mathbf{v}_k^\top, 1) \; F(\mathbf{p}) \begin{pmatrix}\mathbf{x}_k \\ 1\end{pmatrix} = 0$ 
    \EndFor

    \State \# Step 2: Global pose optimization
    \State $Q(\mathbf{f}_k(\mathbf{p}))$ := $\sum_k \tilde{Q}_k(\mathbf{v}_{k,opt})$
    \State $\mathbf{p}$ := argmin $Q(\mathbf{f}_k(\mathbf{p}))$
\EndWhile
    
\end{algorithmic}
\end{algorithm}

%% file: Body/04ImplementationDetails.tex
\section{Implementation Details}

\subsection{Descriptor}
We select the ORB detector \cite{rublee2011orb} as the keypoint detection and description algorithm in this work; this is due to its speed at which it can calculate its Rotated-BRIEF descriptors. During testing, we found that it could calculate the descriptor for every pixel (far enough from the edge to generate a descriptor) in a KITTI frame (roughly 400,000 pixels) in 0.411 seconds on an old Intel Core i7-4790K, achieving single-threaded performance. 

For the ORB descriptor, the orientation of the point is calculated in the detection step, which is then used to rotate the descriptor. Since only the keypoint descriptors are computed, our keypoints are not considered to have rotation in these experiments. The initial matching is still performed using rotated descriptors to minimize the introduction of additional outliers.

All dense descriptors calculated from the lowest level of the image pyramid are used for the experiments. This means that the algorithm is far less robust to scale changes, but this is deemed less important in the test datasets, as the scale is relatively consistent between consecutive frames. 

\subsection{Match culling}

During optimization, there is the possibility that the updated keypoint's position will be located outside of the valid region where a new paraboloid can be calculated. If this happens, our strategy is to invalidate the match and stop considering that keypoint for optimization, keeping the previously recorded loss. This occurs if the keypoint is moved so close to the border that a descriptor cannot be calculated. 

\subsection{Keypoint iterations}
Our solution formulation allows for different methods of iterative keypoint movements during optimization. 
In our experiment the keypoints are iteratively moved in the direction of $\mathbf{v}_{k,opt}$ every time a new relative pose is found, which reduces the total loss. Two different strategies of keypoint iteration are tested:

\begin{enumerate}
    \item Move the keypoint some percentage of the total distance between the current keypoint position and the optimum (e.g. 10\%, 20\%, 30\%). 
    \item Move the keypoint directly onto the current optimum if $\|\mathbf{v}_{k,opt}\| < 1 px$. If $|\mathbf{v}_{k,opt}| >= 1 px$, move the keypoint in the direction towards the optimum by $1px$ magnitude.
\end{enumerate}



\subsection{After optimization iteration}

After the non-linear solver has converged to a solution, there is still no guarantee that all the keypoints are exactly on the epipolar line. To solve this, we find the optimal displacement for every keypoint for the final optimal pose configuration and move all points directly to the optimal position $\mathbf{y}_{k,opt} = \mathbf{y}_k + \mathbf{v}_{k,opt}$ on the respective epipolar line. 

For the sake of the experiments, instead of moving the keypoints to their optimal position, we would rather revert them to their original position, such that the motion estimation for each frame pair is independent from each other.


%% file: Body/05Experiments.tex
\section{Experiments}

\subsection{Evaluation metrics}
\begin{itemize}
    \item \textbf{Relative Rotation Error (Rodrigues angle):} The relative rotation error $\mathbf{R}_{err}$ between two frames (estimated $R_{est}$ and ground truth $R_{gt}$) can be interpreted as a single rotation $\rho$ around an axis $\mathbf{n}$ by using the Rodrigues formula. The rotation magnitude $\rho$ of $\mathbf{R}_{err}$ serves as a measure of deviation from the ground truth.
    \item \textbf{Relative Translation Error:} The experiments are carried out with a monocular system, and consequently, the scale cannot be measured; therefore, only the angle between the ground truth $t_{gt}$ and the estimate $t_{est}$ can be inferred about the accuracy of translation. This is done by: 
    \begin{equation}
        t_{err} = arccos \left( \frac{t_{gt} \cdot t_{est}}{|t_{gt}| |t_{est}|} \right)
    \end{equation}
    \item \textbf{Matching residual (Hamming distance):} An additional score is used to evaluate the final matching residual between $d(\mathbf{x}_k)$ and $d(\mathbf{y}_{k} + \mathbf{v}_{k,opt})$.
\end{itemize}

\subsection{Comparison algorithms}
There are very few current, comparable algorithms in the literature that focus on the two-frame relative pose for visual odometry purposes, and therefore, we use the standard re-projection error (RP-E) as a comparison. We implement a two-frame projection error bundle adjustment that optimizes the relative transformation between $I_A$ and $I_B$ by reducing the sum of the distances between the points in $I_B$ and the closest points on the corresponding epipolar line given the transformation parameters $\mathbf{p}$ and the points in $I_A$. 
This can be implemented as a different JET algorithm with the loss function $Q_k(\mathbf{y}_k + \mathbf{v}_k) = r^2$, which is equivalent to a completely circular loss function that centers on $\mathbf{y}_k$.

As an additional baseline, we also report on the error using the 5-point algorithm (5-P) only, which demonstrates the improvement in pose estimation over the initial pose error. 

We also incorporate the rotation results from the original Photometric JET (P-JET) \cite{bradler2017joint} to demonstrate how our methods compare to the original. These results are taken directly from the original paper. 

\subsection{Datasets}

\subsubsection{KITTI:}
The KITTI dataset \cite{Geiger2012CVPR} is the primary dataset used to evaluate our algorithm. This dataset is a much-beloved dataset in the VO/SLAM literature and features several sequences shot from a moving car in an urban landscape in Germany. Therefore, it generally has a lot of feature-rich scenery. The ground truth is obtained through a sensor fusion model that combines GPS and IMU data. Although KITTI comes with a lot of additional sensor data, we perform our tests based solely on grayscale monocular odometry, using only the images captured from a single camera per sequence. The results are then compared with the ground truth pose data from the sequence. All frames of the sequences are used, and the relative motion is calculated only based on adjacent frames. The relative average error between the frame-to-frame predictions of our method and the comparison baselines is shown in Table \ref{tab:Results}. 

\subsubsection{VAROS:}
We also test our sequences on the synthetic VAROS \cite{zwilgmeyer2021varos} dataset. This is a single sequence dataset featuring much more challenging underwater conditions. It is intended to challenge the robustness of VO and SLAM methods with dynamic and low-light scenarios. 
As it is synthetic, it has an objectively true ground truth not influenced by any possible sensor error that might occur in datasets captured in real-world scenarios. The average error between the frame-to-frame predictions of our method and the comparison baselines is shown in Table \ref{tab:Results3} along with an ablation study on the impact of different-sized sampling radii for the paraboloids.

\begin{table*}[htbp]
\caption{Results of the D-JET original P-JET\cite{bradler2017joint}, re-projection error (RP-E) and the 5-point algorithm baseline (5-P) on the KITTI dataset.}
\begin{center}
\resizebox{\textwidth}{!}{\begin{tabular}{|c||c|c|c|c||c|c|c|c||c|c|c|}
\hline
 & \multicolumn{4}{|c||}{Rotation error $|\rho|$ [Deg]} & \multicolumn{4}{|c||}{Translation error [Deg]} &  \multicolumn{3}{|c|}{Avg Hamming distance}\\
\hline
Seq. & D-JET & RP-E & P-JET & 5-P & D-JET & RP-E & P-JET & 5-P & D-JET & RP-E & 5-P \\
\hline 
\hline
00 & 0.094 & 0.074 & 0.096 & 0.119 & 0.075 & 0.070 & 6.749 & 0.044 & 19.42 & 20.46 & 20.98\\
\hline
01 & 0.115 & 0.091 & 0.253 & 0.138 & 0.096 & 0.075 & 7.889 & 0.064 & 18.58 & 19.92  & 19.73 \\
\hline
02 & 0.084 & 0.059 & 0.061 & 0.112 & 0.110 & 0.061 & 1.502 & 0.030 & 21.78 & 21.94 & 23.44 \\
\hline
03 & 0.073 & 0.052 & 0.035 & 0.101 & 0.154 & 0.120 & 0.981 & 0.056 & 17.72 & 18.42 & 20.00 \\
\hline
04 & 0.068 & 0.038 & 0.045 & 0.090 & 0.030 & 0.013 & 0.951 & 0.021 & 19.22 & 20.03 & 21.28 \\
\hline
05 & 0.086 & 0.062 & 0.049 & 0.107 & 0.099 & 0.078 & 1.274 & 0.076 & 17.99 & 18.75 & 20.04 \\
\hline
06 & 0.072 & 0.049 & 0.358 & 0.100 & 0.046 & 0.025 & 10.765 & 0.027 & 21.33 & 22.04 & 23.18 \\
\hline
07 & 0.112 & 0.081 & 0.152 & 0.107 & 0.206 & 0.177 & 28.070 & 0.169 & 16.93 & 17.70 & 18.97 \\
\hline
08 & 0.078 & 0.060 & 0.063 & 0.108 & 0.092 & 0.069 & 9.429 & 0.064 & 20.25 & 20.32 & 21.78 \\
\hline
09 & 0.087 & 0.056 & 0.029 & 0.109 & 0.067 & 0.031 & 0.676 & 0.030 & 20.30 & 21.21 & 22.65 \\
\hline
10 & 0.116 & 0.075 & 0.131 & 0.146 & 0.093 & 0.072 & 1.361 & 0.057 & 19.97 & 20.96 & 22.53 \\
\hline
\end{tabular}}
\label{tab:Results}
\end{center}
\end{table*}

\begin{table*}[htbp]
\caption{Results of the D-JET on the VAROS sequence with different sampling radii.}
\begin{center}
\resizebox{\textwidth}{!}{\begin{tabular}{|c|c|c|c|c|}
\hline
Method & Paraboloid sampling area & Rotation error $|\rho|$ [Deg] & Translation error [Deg] & Avg Hamming distance \\
\hline
\hline
5-P & - & 0.135 & 0.256 & 22.41\\
\hline
RP-E & - & 0.085 & 0.224 & 22.18\\
\hline
DJET & 3x3 & 0.133 & 0.273 & 21.92\\
\hline
DJET & 5x5 & 0.126 & 0.267 & 21.84\\
\hline
DJET & 7x7 & 0.132 & 0.265 & 21.99\\
\hline
DJET & 15x15 & 0.127 & 0.258 & 22.31\\
\hline
\end{tabular}}
\label{tab:Results3}
\end{center}
\end{table*}

%% file: Body/06Discussion.tex
\section{Results and Discussion}
\subsection{KITTI dataset analysis}
First, we compare the results of D-JET against those of the original algorithm P-JET on the KITTI dataset in Table \ref{tab:Results}. The initial goal of this paper was to achieve an accuracy score comparable to that of the P-JET while increasing the robustness to the level of feature-based methods. As shown in Table \ref{tab:Results}, the algorithm achieves some success in this regard. We observe that for the sequences where the P-JET algorithm achieves less than 0.1 degrees of rotation error, the P-JET generally performs somewhat better than the D-JET algorithm, but for sequences where the P-JET algorithm achieves more than 0.1 degrees of rotation error, D-JET vastly outperforms P-JET, producing much more consistent performance numbers across all sequences. We reason that this consistency is attributed to the increased robustness of D-JETs to rapid changes in viewpoint compared to that of the P-JET. This causes the D-JET algorithm to be less affected by the difficulty of scene content.
We also believe that the sequences with simpler scenes (less than 0.1 degree of rotation error) favor the P-JET algorithm as the photometric receptacle is smaller than the descriptor receptacle and, therefore, produces more peaked loss functions that increase the accuracy when extra robustness is not needed.

A comparison of D-JET and P-JET with the baseline re-projection error (RP-E) yields somewhat surprising results. The standard re-projection error outperforms D-JET in all test sequences and outperforms P-JET in all but two sequences. For P-JET, this can possibly be attributed to the fact that P-JET does not perform any outlier detection/rejection with RANSAC \cite{choi1997performance}, but given that the optimization criteria of P-JET are also based on similarity only along the epipolar line, outliers should have minimal effect on the optimization. 

The discrepancy between D-JET, P-JET, and re-projection error has proved quite challenging to conclude, but by inspecting the average Hamming distances of the different tests in table \ref{tab:Results}, and \ref{tab:Results3}, it can be seen that although the Hamming distance is correlated with accuracy for the D-JET algorithm, it is not a perfect optimization criterion. This is especially clear when comparing the Hamming distances of the re-projection error and the D-JET method with those of the rotational and translational errors. The re-projection error has a higher average Hamming distance, yet it still achieves higher accuracy. We believe these points point to the conclusion that descriptor similarity is not an ideal optimization criterion and is too rough an estimation of similarity.

\subsection{VAROS dataset analysis}
As an ablation study and to rule out biases in the KITTI ground truth as a possible error source, we also tested with the VAROS synthetic dataset. These results can be seen in Table \ref{tab:Results3}. Here, we see an even greater difference between the D-JET algorithm and the re-projection algorithm than was present in the KITTI dataset. We believe this is related to the fact that the content of the VAROS sequence is much more uniform in nature. Causing both incorrect keypoint placement and shallow paraboloids, leading to a less precise optimization. We see that this has such a large effect on D-JET that it is barely better than the linear baseline in this scenario.

\subsection{Ablation: Paraboloid sampling size}

Table \ref{tab:Results3} shows an ablation study looking into the different kernel sizes used for sampling of local regions to approximate the paraboloid. The first thing we observe is that it is not very clear what the best sampling radius is, but a value between 5 and 9 seems to be a good choice. However, more interesting is the last two lines in table \ref{tab:Results3}, where the average hamming distance increases with a kernel size of 15 compared to a kernel size of 7, but both the rotation error and translation error decrease. This shows again that although our self-similarity and cross-similarity experiments give promising results, descriptor similarity does not necessarily strictly correlate with optimal keypoint placement.



%% file: Body/07Conclusion.tex
\section{Conclusion}
In this work, we develop an algorithm that uses descriptor similarity as a pose optimization metric. We have shown that although using photometric optimization schemes on geometric feature descriptor distances is an attractive idea, the descriptors are not ideal criteria for pose optimization, but rather a rough estimation of similarity. Proficient in accurately determining data association from discrete points, but not good enough for approximations in continuous space. Our conclusion is that optimizing for Hamming distance does not directly correlate to an improvement in pose, and re-projection error performs better over the sequences we tested, while also being faster.

%% file: Body/Appendix.tex
\section{Appendix}

\subsection{Quadratic fitting}
The formula we are trying to fit in a quadratic fitting is given as:

\begin{equation}
    f(\mathbf{x}) = \mathbf{x}^T \mathbf{A} \mathbf{x}
    \label{eq:quad_fit}
\end{equation}

The matrix $\mathbf{A}$ represents a 3x3 matrix containing the constants of the quadratic form we are trying to estimate, and $\mathbf{x}$ is a homogeneous vector with 3 elements:

\begin{align}
\begin{split}
    \mathbf{A} =
    \begin{pmatrix}
        a_{11} & \frac{a_{12}}{2} & \frac{a_{1}}{2}\\
        \frac{a_{12}}{2} & a_{22} & \frac{a_{2}}{2}\\
        \frac{a_{1}}{2} & \frac{a_{2}}{2} & a_{0}
    \end{pmatrix}, & \; \; \; \;
    \mathbf{x} =
    \begin{pmatrix}
        x_1\\
        x_2\\
        1
    \end{pmatrix}
\end{split}
\end{align}

Equation \ref{eq:quad_fit} can be reconstructed into a linear least squares problem by setting up the following equation for every sample $i$. Let $\mathbf{d}_i$ be generated from the sample coordinates $\mathbf{x} = \begin{pmatrix} x_1 & x_2 & 1 \end{pmatrix}^T$ and $z_i$ be the measured value for that sample then:

\begin{equation}
    z_i = \underbrace{\begin{pmatrix} x_1^2\; & x_2^2\; & x_1x_2\; & x_1\; & x_2\; & 1 \end{pmatrix}}_{\mathbf{d}_i} \begin{pmatrix} a_{11} \\ a_{22} \\ a_{12} \\ a_{1} \\ a_{2} \\ a_{0} \end{pmatrix}
\end{equation}


\begin{equation}
    \mathbf{z} = \mathbf{D} \cdot \mathbf{a}
\end{equation}

This can be solved as a least squares problem of the general linear form.

The A matrix can be reconstructed by rearranging $\mathbf{a}$ into a matrix while exploiting the inherent symmetry of $\mathbf{A}$.